\documentclass[letterpaper, 10 pt, conference]{ieeeconf}

\IEEEoverridecommandlockouts
\usepackage{graphicx}
\usepackage{amsmath}
\usepackage{amssymb}
\usepackage{booktabs}
\usepackage{hyperref}
\usepackage{cite}
\usepackage[utf8]{inputenc}
\usepackage[table]{xcolor}
\usepackage{tikz}
\usetikzlibrary{arrows.meta,positioning,fit,backgrounds,calc}
\usepackage{dblfloatfix}
\usepackage{cuted}
\usepackage{capt-of}
\usepackage{subfloat}

\makeatletter
\providecommand{\@setmarks}{}
\makeatother

\definecolor{bestcell}{HTML}{E8F5E9}
\newcommand{\best}[1]{\cellcolor{bestcell}\textbf{#1}}

\title{\LARGE \bf
SpaceSense-Bench: A Large-Scale Multi-Modal Benchmark for Spacecraft Perception and Pose Estimation
}

\author{Aodi Wu$^{1,2}$, Jianhong Zuo$^{3}$, Zeyuan Zhao$^{1,2}$, Xubo Luo$^{1,2}$, Ruisuo Wang$^{2}$, and Xue Wan$^{2,*}$%
\thanks{$^{1}$University of Chinese Academy of Sciences, Beijing, China}%
\thanks{$^{2}$Technology and Engineering Center for Space Utilization, Chinese Academy of Sciences, Beijing, China}%
\thanks{$^{3}$Nanjing University of Aeronautics and Astronautics, Nanjing, China}%
\thanks{$^{*}$Corresponding author. Emails: \texttt{wuaodi20@mails.ucas.ac.cn}, \texttt{wanxue@csu.ac.cn}}%
}

\begin{document}

\maketitle
\thispagestyle{empty}
\pagestyle{empty}

\begin{strip}
\centering
\includegraphics[width=0.85\textwidth]{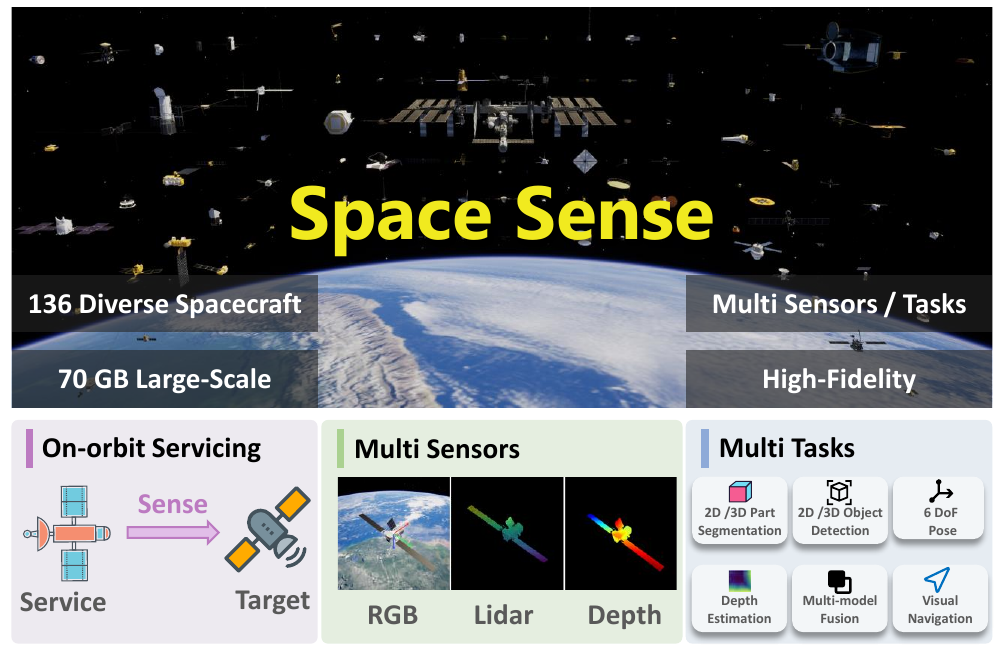}
\captionof{figure}{\textbf{Overview of SpaceSense-Bench.} \textbf{Top:} 136 diverse satellite models rendered in a high-fidelity space environment built with Unreal Engine~5. \textbf{Bottom left:} the on-orbit servicing scenario in which a servicing spacecraft perceives a non-cooperative target. \textbf{Bottom middle:} time-synchronized multi-modal data (RGB image, LiDAR point cloud, and depth map) provided for every frame. \textbf{Bottom right:} the six downstream tasks supported by the benchmark, including 2D/3D part segmentation, object detection, 6-DoF pose estimation, depth estimation, multi-modal fusion, and visual navigation.}
\label{fig:teaser}
\end{strip}

\begin{abstract}
Autonomous space operations such as on-orbit servicing and active debris removal demand robust part-level semantic understanding and precise relative navigation of target spacecraft, yet collecting large-scale real data in orbit remains impractical due to cost and access constraints. Existing synthetic datasets, moreover, suffer from limited target diversity, single-modality sensing, and incomplete ground-truth annotations. We present \textbf{SpaceSense-Bench}, a large-scale multi-modal benchmark for spacecraft perception encompassing 136~satellite models with approximately 70~GB of data. Each frame provides time-synchronized 1024$\times$1024 RGB images, millimeter-precision depth maps, and 256-beam LiDAR point clouds, together with dense 7-class part-level semantic labels at both the pixel and point level as well as accurate 6-DoF pose ground truth. The dataset is generated through a high-fidelity space simulation built in Unreal Engine~5 and a fully automated pipeline covering data acquisition, multi-stage quality control, and conversion to mainstream formats. We benchmark five representative tasks (object detection, 2D semantic segmentation, RGB--LiDAR fusion-based 3D point cloud segmentation, monocular depth estimation, and orientation estimation) and identify two key findings: (i)~perceiving small-scale components (\emph{e.g.}, thrusters and omni-antennas) and generalizing to entirely unseen spacecraft in a zero-shot setting remain critical bottlenecks for current methods, and (ii)~scaling up the number of training satellites yields substantial performance gains on novel targets, underscoring the value of large-scale, diverse datasets for space perception research. The dataset, code, and toolkit are publicly available at \url{https://github.com/wuaodi/SpaceSense-Bench}.
\end{abstract}

\section{INTRODUCTION}

Autonomous space operations, including on-orbit servicing and assembly~\cite{arney2021orbit}, active debris removal~\cite{forshaw2016removedebris,biesbroek2021clearspace}, and proximity rendezvous, are gaining increasing strategic importance as low Earth orbit becomes ever more congested. These missions place demanding perception requirements on the chaser spacecraft: it must achieve part-level semantic understanding of the target, localizing and delineating components such as solar panels, antennas, and thrusters, while simultaneously estimating the target's full 6-DoF relative pose and recovering its 3D structure, all to enable reliable autonomous trajectory planning and manipulation.

However, space perception faces multiple extreme challenges. At the environmental level, orbital illumination varies drastically, from harsh direct sunlight and specular reflections off metallic surfaces, through diffuse Earth Albedo, to near-total darkness in eclipse, creating a high dynamic range that frequently renders RGB-only sensing unreliable and motivates the integration of active sensors such as depth cameras and LiDAR. At the target level, spacecraft encountered during servicing missions span an enormous range of geometries and scales (from 0.27~m CubeSats to 112~m structures like the International Space Station), demanding strong generalization, particularly to entirely unseen targets. Compounding these difficulties, collecting large-scale, multi-modal data with diverse targets in orbit remains impractical, making high-fidelity synthetic datasets the most viable path forward.

In recent years, data-driven deep learning methods have been widely adopted for spacecraft perception~\cite{sharma2019pose,park2024spnv2,crospace6d}, yet their performance hinges critically on the quality and diversity of training data. As summarized in Table~\ref{tab:comparison}, existing spacecraft datasets share three fundamental limitations. First, most datasets contain only 1--16~satellite models, which is insufficient for learning generalizable geometry. Second, none provides time-synchronized RGB, depth, and LiDAR data. Third, the majority lack dense pixel- and point-level part semantics. These shortcomings cause models to overfit to specific satellite appearances and degrade sharply on entirely unseen configurations.

\begin{table}[t]
\caption{\textbf{Comparison with existing spacecraft perception datasets.} Sat: number of spacecraft models. Img: number of frames (SpaceSense-Bench reports sparse / dense). Part Seg classes in parentheses.}
\label{tab:comparison}
\centering
\footnotesize
\setlength{\tabcolsep}{2.2pt}
\begin{tabular}{lrrcccccc}
\toprule
Dataset & Sat & Img & RGB & Depth & LiDAR & Part Seg & 3D Seg & Pose \\
\midrule
SPEED \cite{speed}       & 1   & 15k  & \checkmark &            &            &              &            & \checkmark \\
SPEED+ \cite{speedplus}  & 1   & 70k  & \checkmark &            &            &              &            & \checkmark \\
URSO \cite{urso}         & 2   & 5k   & \checkmark &            &            &              &            & \checkmark \\
Hoang \cite{hoang2021}   & --  & 3k   & \checkmark &            &            & \checkmark (3) &          &            \\
UESD \cite{uesd}         & 33  & 10k  & \checkmark &            &            & \checkmark (5) &          &            \\
SPARK \cite{spark}       & 10  & 150k & \checkmark & \checkmark &            &              &            & \checkmark \\
NCSTP \cite{liu2025large}       & 16  & 200k & \checkmark &            &            & \checkmark (4) &          &            \\
\midrule
\textbf{Ours} & \textbf{136} & \textbf{90k/2M} & \checkmark & \checkmark & \checkmark & \checkmark~\textbf{(7)} & \checkmark & \checkmark \\
\bottomrule
\end{tabular}
\end{table}

To address these limitations, we introduce \textbf{SpaceSense-Bench}, a large-scale multi-modal benchmark encompassing 136~satellite models. Each frame provides time-synchronized 1024$\times$1024 RGB images, millimeter-precision depth maps, and 256-beam LiDAR point clouds, together with dense 7-class part-level semantic labels (pixel and point level) and 6-DoF pose ground truth. A fully automated pipeline built on Unreal Engine~5 and the AirSim~\cite{airsim} plugin handles data acquisition, quality control, and format conversion, enabling cost-effective dataset expansion.

The main contributions of this paper are as follows:
\begin{itemize}
    \item We present SpaceSense-Bench, the first large-scale spacecraft perception benchmark that jointly provides RGB images, depth maps, and LiDAR point clouds, covering 136~satellite models with approximately 70~GB of data.
    \item We supply dense 7-class part-level semantic annotations at both the pixel and point level together with accurate 6-DoF pose ground truth for every frame, automatically converted into YOLO, MMSegmentation, and SemanticKITTI formats to support detection, segmentation, pose estimation, and depth estimation out of the box.
    \item We build a photorealistic space simulation in Unreal Engine~5 and implement a fully automated pipeline for data generation, quality control, and format conversion, enabling low-cost continuous dataset expansion.
    \item We conduct systematic benchmarks across five tasks (2D and 3D semantic segmentation, object detection, depth estimation, and orientation estimation), revealing the challenge of perceiving small-scale components and demonstrating that scaling up training satellites substantially improves zero-shot generalization to unseen targets.
\end{itemize}

\section{RELATED WORK}

\subsection{Spacecraft Datasets and Benchmarks}

Several spacecraft datasets have been proposed in recent years to advance space perception algorithms. SPEED~\cite{speed} and SPEED+~\cite{speedplus} provide synthetic and real images of a single Tango satellite for pose estimation and remain the most widely cited benchmarks in the field. URSO~\cite{urso} renders photorealistic imagery of two spacecraft classes using Unreal Engine. SPARK~\cite{spark} offers approximately 150k RGB and depth images spanning 10~spacecraft and 1~debris class, and is the first public dataset to incorporate a depth sensor. On the part-recognition front, Hoang~et~al.~\cite{hoang2021} construct the first spacecraft part detection and segmentation dataset with 3~component classes, while UESD~\cite{uesd} generates 10,000~images covering 5~part categories across 33~satellite models. More recently, NCSTP~\cite{liu2025large} uses Blender to synthesize 200k~images of 16~satellites, 6~debris types, and 4~space rocks with 4-class part annotations, yet provides only RGB data and does not support pose estimation. As summarized in Table~\ref{tab:comparison}, existing datasets share three critical gaps: none provides time-synchronized RGB--depth--LiDAR triplets, none offers dense pixel-level and point-level part semantics, and none supports 3D point cloud segmentation. SpaceSense-Bench simultaneously addresses all three gaps through its 136-satellite scale, tri-modal sensing, and dense 7-class 2D/3D annotations with 6-DoF pose.

\subsection{Vision-based Spacecraft Semantic Perception}

On-orbit servicing tasks such as robotic grasping of docking rings and solar panel repair demand that perception systems not only detect the target spacecraft but also precisely identify the spatial extent and boundaries of its fine-grained components. Prior work has adapted general-purpose semantic segmentation methods to the spacecraft domain: Zhao~et~al.~\cite{zhao2022_3dsatnet} propose a 3D component segmentation network for non-cooperative spacecraft, Qiu~et~al.~\cite{qiu2022} present an edge-assisted segmentation network and show that DeepLabV3+~\cite{deeplabv3plus} struggles at component boundaries, while Shao~et~al.~\cite{shao2023} introduce a lightweight spatiotemporal segmentation network targeting real-time onboard deployment. Although state-of-the-art architectures such as SegFormer~\cite{segformer} and Mask2Former~\cite{mask2former} have achieved impressive results on natural-image benchmarks, their performance degrades drastically when applied to spacecraft configurations absent from the training set, since existing datasets typically contain only 1--2~satellite models, causing these networks to memorize target-specific textures rather than learn generalizable part geometry. To advance this line of research, a benchmark needs to cover substantially more satellite geometries and enforce strict train--test separation so that models cannot simply memorize target-specific textures. Our dataset provides exactly this setting: 136~satellites with a unified 7-class part taxonomy, evaluated under a zero-shot protocol where training and test spacecraft are entirely non-overlapping.

\subsection{Vision-based Spacecraft Pose Estimation}

Spacecraft 6-DoF pose estimation is a core capability for autonomous proximity navigation. Mainstream approaches include two-stage pipelines based on keypoint regression followed by PnP solving~\cite{sharma2019pose}, end-to-end direct pose regression~\cite{park2024spnv2}, and temporal methods that fuse single-frame and inter-frame estimates~\cite{crospace6d}. As Pauly~et~al.~\cite{murali2023survey} note in their survey, most current methods achieve high accuracy on the single Tango satellite in SPEED~\cite{speed}, yet the SPEC~2021 competition~\cite{spec2021} reveals an order-of-magnitude performance gap when transferring from synthetic to real domains. More critically, this high accuracy is strongly overfitted: models learn the specific geometry of a single target and fail to converge when the satellite configuration changes, \emph{e.g.}, from a compact bus to one with deployed solar arrays or a large dish antenna. Evaluating pose methods across a much wider range of geometries is therefore essential, yet no existing benchmark offers this diversity. Our dataset fills this gap with 136~geometrically distinct configurations and accurate 6-DoF pose ground truth.

\subsection{Multi-modal Perception in Space}

The space environment poses extreme challenges for purely vision-based perception: specular reflections and sharp shadows in sunlit regions, complex illumination from Earth Albedo, and near-total loss of optical texture during eclipse all push RGB sensors close to failure in many operational scenarios. Depth sensors and LiDAR, being independent of ambient illumination, can directly capture geometric structure. In autonomous driving, fusing RGB, depth, and LiDAR has become standard practice for robust perception. In the space domain, however, although SPARK~\cite{spark} introduced RGB-D data and catalyzed early multi-modal recognition studies~\cite{aldahoul2021}, it still lacks a LiDAR modality and does not provide strictly time-synchronized multi-sensor captures. This data gap has severely hindered the exploration of multi-modal fusion for space target perception. By providing time-synchronized RGB images, millimeter-precision depth maps, and 256-beam LiDAR point clouds in every frame, all spatiotemporally aligned, our dataset opens the door to systematic multi-modal fusion research in the space domain.

\section{SPACESENSE-BENCH DATASET}

\begin{figure*}[!t]
\centering
\includegraphics[width=\textwidth]{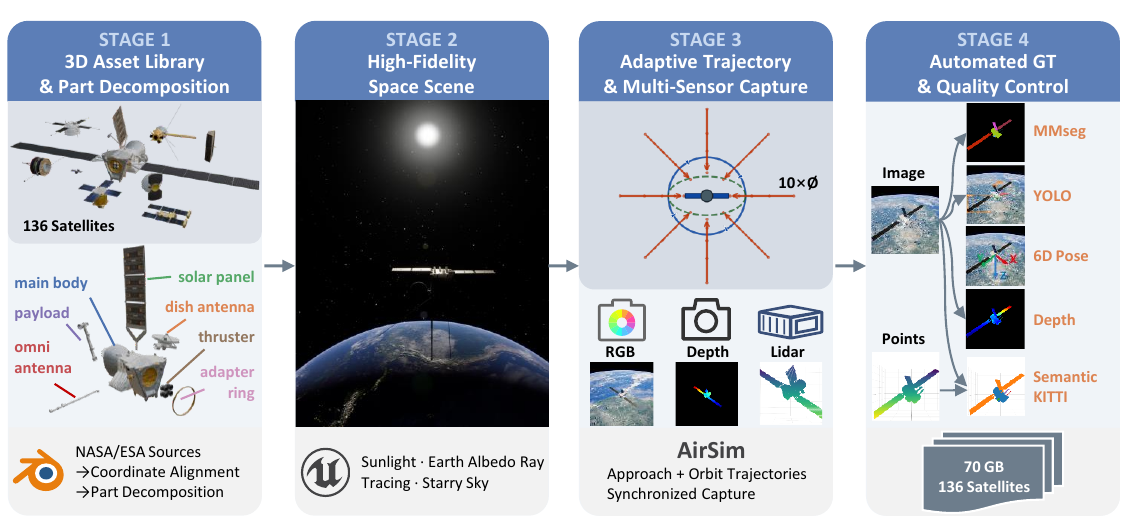}
\caption{\textbf{Overall data collection pipeline of SpaceSense-Bench.} The pipeline consists of four stages: (1) 3D asset library construction and part decomposition, (2) high-fidelity space scene setup, (3) adaptive trajectory planning and multi-sensor synchronized capture, and (4) automated ground truth generation, quality control, and mainstream format export.}
\label{fig:pipeline}
\end{figure*}

\subsection{Dataset Overview}

As illustrated in Fig.~\ref{fig:pipeline}, the construction of SpaceSense-Bench proceeds in four fully automated stages: (1)~3D asset library construction and part-level semantic decomposition in Blender, (2)~high-fidelity space scene setup in Unreal Engine~5, (3)~scale-adaptive trajectory planning with multi-sensor synchronized capture via the AirSim plugin, and (4)~ground-truth generation, quality control, and mainstream format export. The entire pipeline requires no manual intervention, enabling efficient scaling to 136~satellites at low marginal cost.

\begin{figure*}[!t]
\centering
\includegraphics[width=\textwidth]{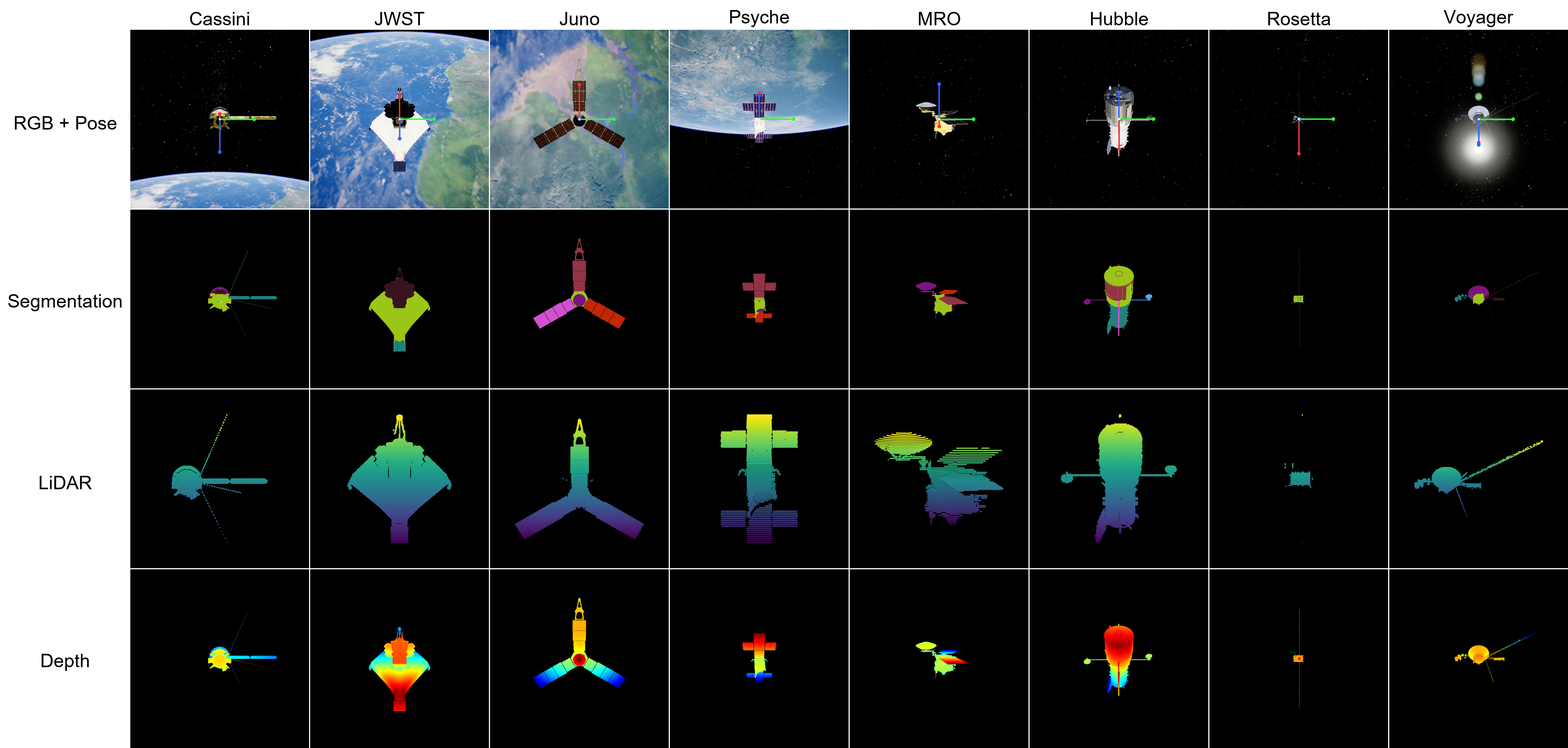}
\caption{\textbf{Multi-modal data and ground truth examples from SpaceSense-Bench.} Each column shows one satellite. From top to bottom: RGB image with 6D pose axes overlay, seven-class semantic segmentation mask, LiDAR point cloud with per-point semantic labels, and colorized depth map.}
\label{fig:samples}
\end{figure*}

The resulting dataset comprises 136~satellite models of diverse appearance, function, and scale (maximum diameters from 0.27~m to 112~m), totaling approximately 70~GB. Under the current \emph{sparse} sampling strategy, the dataset contains 90k~synchronously captured frames; by switching to \emph{dense} capture with finer angular and distance steps, the fully automated pipeline can scale to over 2M~frames without additional manual effort. As shown in Fig.~\ref{fig:samples}, each frame simultaneously provides a 1024$\times$1024 RGB image, a millimeter-precision depth map, a 256-beam LiDAR point cloud, dense 7-class part-level semantic labels at both the pixel and point level, and 6-DoF pose ground truth in the camera coordinate system. All data are automatically converted into YOLO, MMSegmentation, and SemanticKITTI formats for 2D detection, 2D segmentation, and 3D point cloud segmentation, respectively, ready for immediate use. All experiments in this paper are conducted on the sparse-sampled 90k-frame subset.

\subsection{3D Asset Library and Taxonomy}

We aggregate satellite 3D models from publicly available platforms including NASA 3D Resources, NASA Eyes on the Solar System, and ESA Science Satellite Fleet. After deduplication and curation, we obtain 136~satellites spanning communication satellites, science probes, planetary explorers, and space telescopes, with physical dimensions ranging from 0.27~m (CubeSat-class) to 112~m (ISS-class). To ensure consistency within the simulation, we use Blender to standardize every model's coordinate frame ($+X$ velocity, $+Y$ starboard, $+Z$ nadir) and export all assets in textured FBX format.

Unlike conventional workflows that annotate each 2D frame by hand, we perform instance-level part decomposition directly on the 3D mesh in Blender, assigning a unique Stencil ID to each component. This design allows the rendering engine to produce pixel-precise semantic masks automatically for an unlimited number of viewpoints, completely eliminating manual labeling cost. The part taxonomy is designed by jointly considering the structural commonality across satellite platforms and the functional importance for on-orbit servicing. Despite their varied appearances, the vast majority of satellites share a common set of functional modules: structural bus, solar power generation, communication, mission payload, propulsion, and docking interface (see Stage~1 in Fig.~\ref{fig:pipeline}). Accordingly, we define seven semantic part classes: (1)~\texttt{main\_body} (primary structural bus), (2)~\texttt{solar\_panel} (solar arrays and support structures), (3)~\texttt{dish\_antenna} (parabolic high-gain antennas), (4)~\texttt{omni\_antenna} (rod-shaped or helical omnidirectional antennas), (5)~\texttt{payload} (mission instruments, science sensors, and booms), (6)~\texttt{thruster} (propulsion nozzles), and (7)~\texttt{adapter\_ring} (launch vehicle adapter ring).

\subsection{High-Fidelity Simulation Environment}

The real space environment exhibits three primary illumination regimes: direct sunlight in sun-lit regions, diffuse reflected light from Earth's surface (Earth Albedo), and near-total darkness during eclipse, collectively posing an extreme high-dynamic-range challenge for visual perception. To faithfully reproduce these conditions, we configure a directional light in Unreal Engine~5 to simulate solar irradiance, model the Earth with high-resolution surface and cloud-layer textures using the SkyAtmosphere system and SkyLight for ambient illumination, enable hardware ray tracing for physically accurate specular reflections off metallic surfaces and sharp cast shadows, and construct a deep-space backdrop based on real star catalogs.

On the virtual chaser spacecraft, we mount three spatially co-located sensors: an RGB camera (1024$\times$1024 resolution, 50$^\circ$ field of view), a perspective depth camera (same resolution, millimeter precision), and a 256-beam LiDAR (40$^\circ$ field of view). Because all sensors capture data from the same position, the resulting multi-modal outputs are strictly aligned in both space and time, providing a reliable foundation for multi-modal fusion research.

\subsection{Automated Data Generation and Quality Control}

During proximity operations, the chaser spacecraft gradually approaches the target, causing the target's apparent scale to vary by several orders of magnitude. To cover this full range, we design two trajectory families for each satellite: \emph{approach} and \emph{orbit}. Approach trajectories start at 10$\times$ the target's maximum diameter and advance toward 1$\times$ diameter with geometrically decaying steps, each covering 10\% of the current distance to ensure uniform target growth in the field of view, along 22~distinct approach directions spanning 6~axial and 16~diagonal combinations. Orbit trajectories circle the target at 2$\times$ diameter with a 1$^\circ$ angular step over a full 360$^\circ$ sweep across 5~orbital planes. Because both step sizes and radii are adaptively scaled by the target's maximum diameter, satellites ranging from a 0.27~m CubeSat to the 112~m ISS all receive appropriate sampling density.

The entire acquisition pipeline operates in a fully automated, one-click fashion (Stages~3--4 in Fig.~\ref{fig:pipeline}): for each of the 136~satellites, the system sequentially performs trajectory generation, model swapping, Stencil ID assignment, AirSim multi-sensor synchronized capture, and ground-truth export. At every frame, the pipeline simultaneously records an RGB image, a pixel-level 7-class semantic mask derived from the Custom Depth Stencil, a millimeter-precision depth map, a LiDAR point cloud with per-point semantic labels projected from the mask, and the target's 6-DoF pose. Automated scripts then batch-convert all data into YOLO, MMSegmentation, and SemanticKITTI formats. A multi-level quality control stage filters anomalous frames by checking label validity, depth coverage, point cloud integrity, and cross-modal timestamp consistency; failing frames are discarded and re-acquired.

\subsection{Dataset Statistics and Evaluation Protocol}

\begin{figure*}[thpb]
\centering
\includegraphics[width=\textwidth]{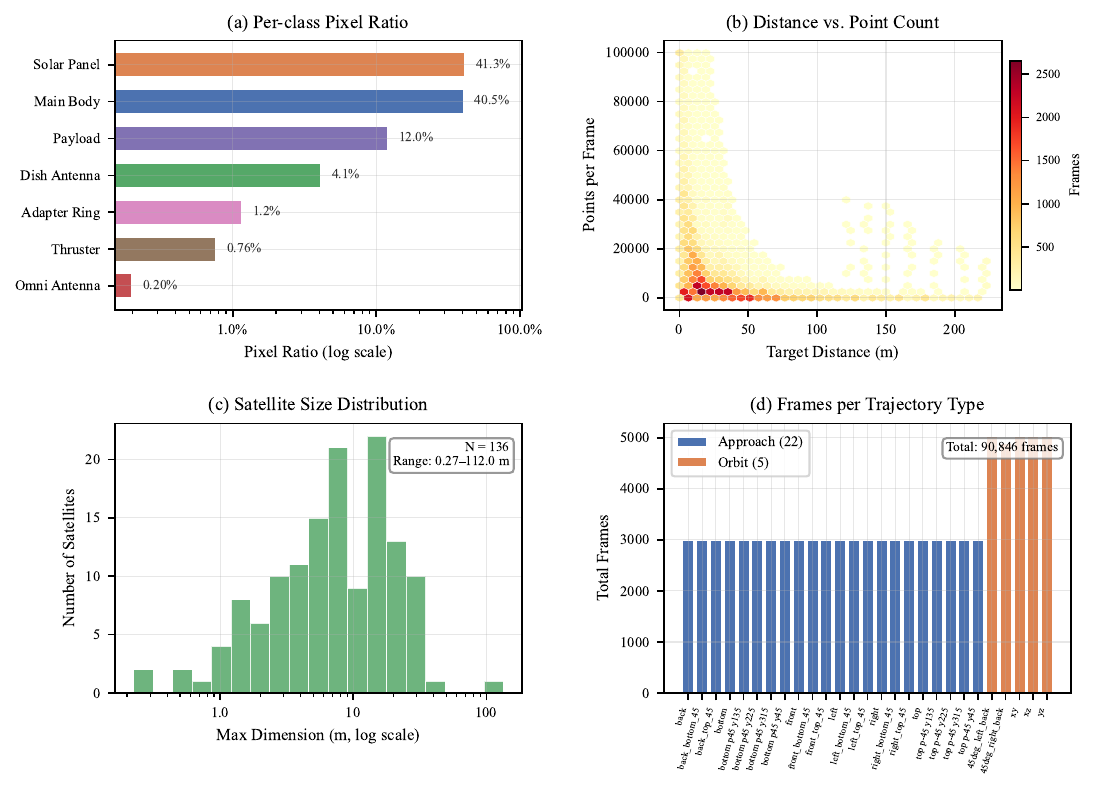}
\caption{\textbf{Dataset statistics.} (a)~Per-class foreground pixel ratio showing a long-tail distribution. (b)~LiDAR point count versus target distance. (c)~Satellite maximum dimension distribution (log scale). (d)~Frame count per trajectory type for 22 approach and 5 orbit trajectories.}
\label{fig:distribution}
\end{figure*}

Under the sparse sampling strategy, SpaceSense-Bench contains 90k~synchronously captured multi-modal frames (RGB, semantic segmentation masks, depth maps, LiDAR point clouds, and 6-DoF pose), covering 136~satellites across 27~trajectory types (22~approach + 5~orbit), with satellite maximum dimensions spanning 0.27~m to 112~m (Fig.~\ref{fig:distribution}c). As shown in Fig.~\ref{fig:distribution}(a), the per-class pixel distribution exhibits a pronounced long-tail pattern. \texttt{solar\_panel} and \texttt{main\_body} each account for approximately 41\% of foreground pixels, together exceeding 80\%, while \texttt{thruster} (0.76\%), \texttt{adapter\_ring} (1.2\%), and \texttt{omni\_antenna} (0.20\%) occupy only marginal fractions, posing a severe small-object perception challenge. Fig.~\ref{fig:distribution}(b) illustrates the relationship between target distance and per-frame LiDAR point count: at close range ($<$30~m) each frame captures tens of thousands of points, whereas beyond 100~m the count drops to a few hundred, reflecting the multi-scale difficulty inherent in approach scenarios. Fig.~\ref{fig:distribution}(d) confirms that frame counts are evenly distributed across trajectory types ($\sim$3{,}000~frames per approach type, $\sim$4{,}900 per orbit type), ensuring balanced training and evaluation.

To rigorously evaluate generalization to unseen spacecraft, we adopt an out-of-distribution (OOD) protocol, partitioning the 136~satellites into 117 for training, 5 for validation, and 14 for testing, with zero model overlap across splits. This zero-shot setup ensures that every test satellite is entirely absent during training, forcing algorithms to learn generalizable geometric cues of spacecraft parts, such as the rectangular array pattern of solar panels or the parabolic profile of dish antennas, rather than memorizing the texture of specific targets.

\section{BENCHMARK EXPERIMENTS}

To comprehensively evaluate SpaceSense-Bench, we conduct systematic experiments on five tasks: 2D semantic segmentation, object detection, 3D point cloud segmentation, monocular depth estimation, and orientation estimation. The first three tasks are evaluated with models trained on the dataset, while the latter two employ pre-trained vision foundation models in a zero-shot setting to probe cross-domain generalization. We further perform a data scaling ablation to quantify the impact of training set size on zero-shot generalization to unseen spacecraft.

\subsection{Experimental Setup}

We benchmark five tasks with representative baselines. For \textbf{2D semantic segmentation}, we train FCN~\cite{fcn} (ResNet-50), DeepLabV3+~\cite{deeplabv3plus} (ResNet-50), SegFormer~\cite{segformer} (MiT-B3), and Mask2Former~\cite{mask2former} (Swin-Base), all initialized from ImageNet pre-trained weights, and report mean Intersection-over-Union (mIoU) and overall pixel accuracy (aAcc). For \textbf{object detection}, we train YOLO26~\cite{yolo26} at five scales (n/s/m/l/x) and report mAP@0.5 and mAP@0.5:0.95. For \textbf{3D point cloud segmentation}, we train PMFNet~\cite{pmfnet} (ResNet-34) which fuses RGB and LiDAR, and report mIoU and frequency-weighted IoU (fwIoU). For \textbf{monocular depth estimation}, we evaluate Depth Anything V2~\cite{depthanythingv2} (ViT-S/B/L) in a zero-shot setting with least-squares affine alignment on the satellite region, reporting AbsRel and Spearman rank correlation. For \textbf{orientation estimation}, we evaluate Orient Anything~\cite{orientanything} (DINOv2-Large) zero-shot, reporting Mean Axis Angular Error (MAAE).

All 2D segmentation models are trained at 640$\times$640 resolution for 50 epochs; YOLO26 is trained for 100 epochs; PMFNet uses a learning rate of 5e-5. All training is conducted on NVIDIA H200 GPUs.

\subsection{Main Results and Analysis}

\begin{table*}[t]
\caption{\textbf{Benchmark results on the 14-satellite zero-shot test set.} Best results in \best{bold}. Task-specific metrics are detailed in each sub-table header.}
\label{tab:main_results}
\centering
\footnotesize
\setlength{\tabcolsep}{3.0pt}
\begin{tabular}{llccccccccc}
\toprule
\rowcolor{gray!12}
\multicolumn{11}{l}{\textit{(a) 2D Semantic Segmentation --- per-class IoU (\%)}} \\
\cmidrule(lr){1-11}
Model & Backbone & aAcc & mIoU & \rotatebox{60}{body} & \rotatebox{60}{solar} & \rotatebox{60}{dish} & \rotatebox{60}{omni} & \rotatebox{60}{payload} & \rotatebox{60}{thruster} & \rotatebox{60}{adapter} \\
\cmidrule(lr){1-11}
FCN~\cite{fcn} & ResNet-50 & 99.16 & 41.30 & 68.1 & 82.9 & 30.7 & 1.0 & 19.0 & 12.8 & 16.1 \\
DeepLabV3+~\cite{deeplabv3plus} & ResNet-50 & 99.19 & 43.70 & 68.1 & 83.2 & 38.2 & 1.0 & 21.5 & 17.7 & 20.2 \\
SegFormer~\cite{segformer} & MiT-B3 & 99.27 & 45.14 & 71.2 & 87.6 & \best{39.7} & \best{2.9} & \best{22.6} & 20.1 & 17.3 \\
Mask2Former~\cite{mask2former} & Swin-B & \best{99.28} & \best{45.63} & \best{71.4} & \best{88.6} & 26.3 & 1.9 & 19.1 & \best{24.6} & \best{33.3} \\
\midrule[0.06em]
\addlinespace[2pt]
\rowcolor{gray!12}
\multicolumn{11}{l}{\textit{(b) Object Detection (YOLO26~\cite{yolo26}) --- per-class AP@0.5 (\%)}} \\
\cmidrule(lr){1-11}
Model & Scale & Prec. & mAP50 & \rotatebox{60}{body} & \rotatebox{60}{solar} & \rotatebox{60}{dish} & \rotatebox{60}{omni} & \rotatebox{60}{payload} & \rotatebox{60}{thruster} & \rotatebox{60}{adapter} \\
\cmidrule(lr){1-11}
YOLO26 & Nano & 48.8 & 33.9 & 89.0 & 78.0 & 19.1 & 3.6 & 8.6 & 8.4 & 30.5 \\
YOLO26 & Small & 53.4 & 37.1 & 89.4 & 80.4 & 19.7 & 7.9 & 8.8 & 10.8 & 42.8 \\
YOLO26 & Medium & 54.6 & 39.0 & \best{91.3} & 82.3 & 21.6 & 6.4 & 10.0 & 16.5 & 45.1 \\
YOLO26 & Large & 52.6 & 39.5 & 90.5 & 81.9 & \best{27.1} & 6.0 & \best{10.8} & 15.3 & 45.2 \\
YOLO26 & XLarge & \best{56.1} & \best{41.3} & 91.0 & \best{82.5} & 23.7 & \best{8.0} & 9.1 & \best{23.3} & \best{51.6} \\
\midrule[0.06em]
\addlinespace[2pt]
\rowcolor{gray!12}
\multicolumn{11}{l}{\textit{(c) 3D Point Cloud Segmentation (PMFNet~\cite{pmfnet}, ResNet-34, RGB+LiDAR fusion) --- per-class IoU (\%)}} \\
\cmidrule(lr){1-11}
Model & Backbone & mAcc & mIoU & \rotatebox{60}{body} & \rotatebox{60}{solar} & \rotatebox{60}{dish} & \rotatebox{60}{omni} & \rotatebox{60}{payload} & \rotatebox{60}{thruster} & \rotatebox{60}{adapter} \\
\cmidrule(lr){1-11}
PMFNet & ResNet-34 & 57.5 & 42.4 & 68.8 & 85.8 & 51.7 & 8.9 & 21.9 & 25.2 & 34.2 \\
\midrule[0.06em]
\addlinespace[2pt]
\rowcolor{gray!12}
\multicolumn{11}{l}{\textit{(d) Monocular Depth Estimation (Depth Anything V2~\cite{depthanythingv2}, zero-shot + affine alignment)}} \\
\cmidrule(lr){1-11}
Model & Backbone & \multicolumn{2}{c}{AbsRel$\downarrow$} & SqRel$\downarrow$ & \multicolumn{2}{c}{RMSE(m)$\downarrow$} & RMSElog$\downarrow$ & $\delta<$1.25$\uparrow$ & \multicolumn{2}{c}{Spearman$\uparrow$} \\
\cmidrule(lr){1-11}
DA-V2 & ViT-S & \multicolumn{2}{c}{0.0236} & 0.0317 & \multicolumn{2}{c}{0.747} & 0.0319 & 99.77\% & \multicolumn{2}{c}{0.555} \\
DA-V2 & ViT-B & \multicolumn{2}{c}{0.0227} & 0.0304 & \multicolumn{2}{c}{\best{0.746}} & 0.0312 & 99.77\% & \multicolumn{2}{c}{0.578} \\
DA-V2 & ViT-L & \multicolumn{2}{c}{\best{0.0223}} & \best{0.0304} & \multicolumn{2}{c}{0.757} & \best{0.0307} & \best{99.77\%} & \multicolumn{2}{c}{\best{0.602}} \\
\midrule[0.06em]
\addlinespace[2pt]
\rowcolor{gray!12}
\multicolumn{11}{l}{\textit{(e) Orientation Estimation (Orient Anything~\cite{orientanything}, DINOv2-Large, zero-shot)}} \\
\cmidrule(lr){1-11}
Model & Backbone & \multicolumn{2}{c}{MAAE$\downarrow$} & \multicolumn{2}{c}{Median$\downarrow$} & $<$10$^\circ$$\uparrow$ & $<$20$^\circ$$\uparrow$ & $<$30$^\circ$$\uparrow$ & \multicolumn{2}{c}{$<$45$^\circ$$\uparrow$} \\
\cmidrule(lr){1-11}
Orient-Any. & DINOv2-L & \multicolumn{2}{c}{12.75$^\circ$} & \multicolumn{2}{c}{10.56$^\circ$} & 53.7\% & 78.2\% & 91.7\% & \multicolumn{2}{c}{98.8\%} \\
\bottomrule
\end{tabular}
\end{table*}

Table~\ref{tab:main_results} reports quantitative results on the test set, where all 14~satellites are entirely absent from training. For 2D semantic segmentation, Mask2Former achieves the best overall mIoU of 45.63\%, closely followed by SegFormer (45.14\%), yet no method exceeds 46\%, indicating that even state-of-the-art architectures face significant challenges when generalizing to unseen spacecraft configurations. For 3D point cloud segmentation, PMFNet attains 42.4\% mIoU via RGB--LiDAR fusion, comparable to the 2D counterparts.

A pronounced long-tail effect is evident across all tasks. \texttt{solar\_panel} and \texttt{main\_body} consistently exceed 68\% IoU, while \texttt{omni\_antenna}, \texttt{thruster}, and \texttt{adapter\_ring} remain below 35\% IoU even for the best-performing method. These components occupy tiny pixel fractions (e.g., \texttt{omni\_antenna} accounts for only 0.20\% of foreground pixels, cf.\ Fig.~\ref{fig:distribution}a), exhibit high morphological diversity, and become indistinguishable from background at long range, making small-component recognition a core bottleneck.

For zero-shot foundation model evaluation, Depth Anything V2 (ViT-L) achieves AbsRel\,=\,0.022 and $\delta<$1.25 accuracy of 99.77\% on the satellite region after affine alignment; however, its Spearman rank correlation is only 0.602, revealing limited capacity for relative depth ordering on metallic space surfaces. Orient Anything attains a mean MAAE of 12.75$^\circ$ with 78.2\% of frames below 20$^\circ$, though performance varies widely, from 7.8$^\circ$ on structurally simple satellites (GRAIL) to 22.1$^\circ$ on compact, symmetric targets (LADEE).

\subsection{Data Scaling Analysis}

\begin{figure}[t]
\centering
\includegraphics[width=\columnwidth]{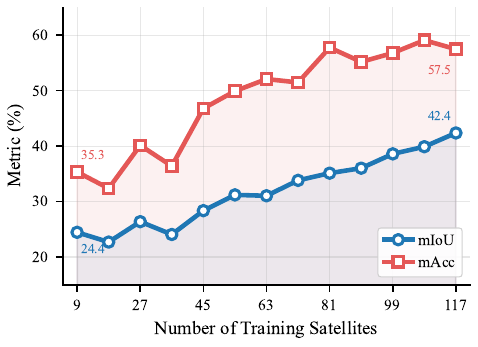}
\caption{\textbf{Effect of training set size on zero-shot generalization.} mIoU and mAcc of PMFNet on the 14-satellite test set as the number of training satellites increases from $\sim$9 to 117.}
\label{fig:scaleup}
\end{figure}

To quantify the impact of dataset scale on zero-shot generalization, we partition the 117~training satellites into 13~equal groups and train PMFNet with progressively more groups while keeping the 14-satellite test set fixed (Fig.~\ref{fig:scaleup}). Both mIoU and mAcc exhibit a clear overall upward trend: mIoU increases from 24.4\% (9 satellites) to 42.4\% (all 117), a 73\% relative gain, while mAcc rises from 35.3\% to 57.5\%, a 63\% relative gain. The growth shows diminishing but non-saturating returns as more satellites are added. These results strongly validate that large-scale, diverse spacecraft datasets are essential for zero-shot generalization, and confirm the value of SpaceSense-Bench's 136-satellite scale.

\section{CONCLUSIONS}

We have presented SpaceSense-Bench, a large-scale multi-modal benchmark covering 136~satellite models with synchronized RGB, depth, and LiDAR data as well as dense 7-class part-level annotations and 6-DoF pose ground truth. Experiments across five perception tasks expose two persistent challenges. First, small-scale components such as \texttt{omni\_antenna}, \texttt{thruster}, and \texttt{adapter\_ring} remain the primary failure mode, staying below 35\% IoU even with the strongest segmentation model, largely due to their tiny pixel footprint ($<$2\% of foreground) and high morphological variation. Second, the data scaling study confirms that adding more satellite geometries continues to improve zero-shot generalization without saturating; mIoU rises 73\% as training satellites grow from 9 to 117, suggesting that further expansion of the model library is a promising path toward robust space perception.

Beyond the five tasks benchmarked in this paper, the multi-modal data and dense annotations in SpaceSense-Bench readily support a broader spectrum of research directions, including 3D reconstruction of spacecraft from multi-view RGB and depth, point-cloud-conditioned image synthesis and cross-modal generation (LiDAR-to-RGB, depth-to-RGB). In future work, we plan to leverage recent advances in 3D generative models to synthesize novel satellite geometries, further scaling the diversity and volume of training data. We also plan to validate sim-to-real transferability using laboratory testbed imagery and real on-orbit data.



\section*{ACKNOWLEDGMENT}
This work was supported by the National Key Research and Development Program of China (Grant No.\ 2024YFB3909300). 
\bibliographystyle{IEEEtran}
\bibliography{references}

\end{document}